%% file: main.tex
\documentclass[manuscript,screen,preprint]{acmart}

\AtBeginDocument{%
  }

\setcopyright{acmlicensed}
\copyrightyear{2025}
\acmYear{2025}
\acmDOI{XXXXXXX.XXXXXXX}

\usepackage{amsmath,amssymb,amsfonts}
\usepackage{algorithmic}
\usepackage{subcaption}
\usepackage{caption}
\usepackage{tabularx}
\usepackage{graphicx}
\usepackage{textcomp}
\usepackage{booktabs}
\usepackage{hyperref}
\usepackage{xcolor}
\usepackage{cleveref}
\usepackage{glossaries}

\newcommand{\ind}{\perp\!\!\!\!\perp} 

\acmISBN{978-1-4503-XXXX-X/2018/06}

\begin{document}
\input{glossaries}
\title{Probabilistic Runtime Verification, Evaluation and Risk Assessment of Visual Deep Learning Systems}

\author{Birk Torpmann-Hagen} \authornote{Also affiliated with SimulaMet}
\affiliation{%
  \institution{UiT: The Arctic University of Norway}
  \city{Tromsø}
  \country{Norway}}
\email{birk.s.torpmannhagen@uit.no}

\author{Pål Halvorsen}\authornote{Also affiliated with Oslo Metropolitan University, Norway}
\affiliation{%
  \institution{SimulaMet}
  \city{Oslo}
  \country{Norway}
}

\author{Michael A. Riegler}
\affiliation{%
 \institution{Simula Research Laboratory}
 \city{Oslo}
 \country{Norway}}

\author{Dag Johansen}
\affiliation{%
  \institution{UiT: The Arctic University of Norway}
  \city{Tromsø}
  \country{Norway}}
\renewcommand{\shortauthors}{Torpmann-Hagen et al.}
\begin{abstract}
Despite achieving excellent performance on benchmarks, deep neural networks often underperform in real-world deployment due to sensitivity to minor, often imperceptible shifts in input data, known as distributional shifts. These shifts are common in practical scenarios but are rarely accounted for during evaluation, leading to inflated performance metrics. To address this gap, we propose a novel methodology for the verification, evaluation, and risk assessment of deep learning systems. Our approach explicitly models the incidence of distributional shifts at runtime by estimating their probability from outputs of out-of-distribution detectors. We combine these estimates with conditional probabilities of network correctness, structuring them in a binary tree. By traversing this tree, we can compute credible and precise estimates of network accuracy. We assess our approach on five different datasets, with which we simulate deployment conditions characterized by differing frequencies of distributional shift. Our approach consistently outperforms conventional evaluation, with accuracy estimation errors typically ranging between 0.01 and 0.1. We further showcase the potential of our approach on a medical segmentation benchmark, wherein we apply our methods towards risk assessment by associating costs with tree nodes, informing cost-benefit analyses and value-judgments. Ultimately, our approach offers a robust framework for improving the reliability and trustworthiness of deep learning systems, particularly in safety-critical applications, by providing more accurate performance estimates and actionable risk assessments.
\end{abstract}

\received{20 February 2007}
\received[revised]{12 March 2009}
\received[accepted]{5 June 2009}

\keywords{
Deep learning, Evaluation, Probabilistic Risk Assessment, runtime verification}

\maketitle
\section{Introduction}

\glspl{dnn} are often considered the state of the art for a wide variety of complex tasks and across a wide variety of domains, consistently scoring highly on benchmarks. However, recent work has shown that these results may not necessarily be indicative of real-world performance~\cite{large-scale-ml-deployment}. Often, there is a considerable gap in the values of the evaluation metrics computed during development, and the actual performance observed in deployment conditions~\cite{endocv2021_review,damour2020underspecification, large-scale-ml-deployment}. This can be attributed to the sensitivity \glspl{dnn} exhibit with respect to certain changes in the nature of the input data that are not well accounted for in the training data --- i.e. \textit{distributional shifts}. This sensitivity can in turn be attributed to the fact that neural networks are generally trained in an underspecified manner, i.e. over-parametrized with respect to the quantity of training data~\cite{damour2020underspecification}. This means that a given training process can yield networks that encode significantly differing interpretations of the data, and typically interpretations that favour dataset-specific covariates with strong predictive power that nevertheless are spurious with respect to causality~\cite{shortcut_learning, cleverhans}. This has been studied extensively across several medical domain case studies, therein for polyp segmentation~\cite{endocv2021_review}, x-ray based pneumomia detection~\cite{pneumonia}, and in the context of hidden stratification in various medical tasks~\cite{skin_shortcut, damour2020underspecification}, and more~\cite{hidden_stratification}. Beyond the medical domain, it has also been shown that \glspl{dnn} trained on ImageNet generalize poorly to ImageNet-like datasets~\cite{imagenet_gen}, with similar behaviour being observed for the CIFAR datasets~\cite{cifar10_generalizability}. While not yet as extensively studied, similar behaviour has been observed in state-of-the-art large language models~\cite{llm_genfailure, llm_generalization}. 

This has significant implications for the outcomes of \glspl{dls}. The poor generalization capability of neural networks, coupled with their black-box nature, means that \glspl{dls} are liable to yield incorrect predictions during deployment for data characterized by distributional shift. This can for instance adversely affect safety-critical systems, such as autonomous vehicles, where it is precisely poorly represented data --- e.g. accidents and situations leading up to accidents --- that are associated with the highest risk~\cite{amodei2016concrete}. Distributional shifts may in a similar manner adversely effect fairness, as poorly represented covariates can in certain domains entail a high degree of social salience. Even discounting outcomes of immediately obvious significance, prediction errors that are unaccounted for will at the very least undermine the functionality of the system as a whole and thus potentially eliminate any benefit that motivated the implementation of a \gls{dnn} in the first place. 

In addition to the concerns posed by a network silently yielding incorrect predictions, there is also the issue that the accuracy estimates used to justify deployment often significantly over-estimate the true level of performance the network attains during deployment. The current praxis for evaluating neural networks simply considers an unseen partition of the training data, e.g. through a holdout set or cross-validation~\cite{challenges_deploying}. The incidence of distributional shifts and their frequency is thus not typically accounted for in evaluations. This neglects the dynamic nature inherent to real-world deployment; the incidence rate of distributional shifts is not only unlikely to be sufficiently accounted for by a hold-out test set, but it is also liable to evolve over time during deployment. While there has been some work towards developing more robust evaluation, notably utilizing synthetic data generation~\cite{deepxplore}, the efficacy of these methods has been contested~\cite{testing_ml}. Regardless, these methods implicitly encode a number of assumptions as part of test case generation. Notably, it presupposes a set incidence rate of distributional shift, a static data environment, and that the generated test cases represent a uniform sampling of data under deployment conditions. We contend that these assumptions rarely hold, and that sufficient characterization of network behaviour requires continuous online monitoring, both in order to account for the diversity of deployment data and in order to account for any drifts in the distribution over time. 

To this end, we develop a comprehensive methodology for runtime assurance, encompassing runtime verification, runtime evaluation, and runtime risk assessment. Our approach is based on \gls{pra}, a methodology often used in other high stakes engineering disciplines, which involves identifying events that can adversely affect a given system and estimating their probability. Central to our method is the construction of an event tree, which maps out the potential outcomes for a \glspl{dnn} along with the associated conditional probabilities. Our primary contribution in this regard lies in the observation that the probability of root events --- e.g. distributional shift --- can be computed at runtime using a appropriate detector --- i.e. a \gls{ood} detector --- coupled with a model of its uncertainty. By traversing the resulting event tree, one can estimate the expected network accuracy over a given time horizon, for which we attain average errors ranging between \(0.02\) and \(0.1\) across five benchmarks. We further contextualize the practical significance of our work on a polyp segmentation case study, for which we illustrate how our methods can be leveraged towards runtime verification, risk characterization, and cost-benefit analyses. 

We summarize our contributions as follows:
\begin{itemize}
    \item We introduce a methodology for the runtime probabilistic evaluation, verification, and risk assessment of \glspl{dnn}.
    \item We show that this framework is capable of accurately estimating network accuracy on deployment data without the need for labels, based on an analysis of five different datasets.
    \item We demonstrate how our framework can be leveraged to yield estimations of risk by associating outcomes with a corresponding cost, and assess these estimations on a polyp segmentation dataset. 
    \item We discuss further extensions of this framework and how it may be applied to characterize the incidence of more cross-cutting system failures such as fairness and safety violations. 
\end{itemize}
All code and data used as part of this work are available at our \href{https://github.com/BirkTorpmannHagen/DeepLearningPRA}{GitHub Repository}. 
The remainder of this work is structured as follows: In \Cref{ch:rw}, we outline related work on uncertainty quantification and distributional shift detection for \glspl{dnn}. \Cref{ch:framework} outlines our framework, including the implementation of likelihood estimators. In \Cref{ch:methodology} we outline the methodology we have utilized in order to assess our framework, the results of which are shown in \Cref{ch:results}. We discuss the implications of our work as well as some potential extensions to our framework in \Cref{ch:discussion}, and finally, we conclude our work in \Cref{ch:conclusion}.

\section{Related Work}\label{ch:rw}

In this section, we outline related work in the field. In particular, we further discuss the prevalence of generalization failure and its effects, how current approaches to the assessment of \gls{dnn} fail to capture these shortcomings, and outline related work on \gls{ood} detectors. 

\subsection{Generalization Failure}
\glspl{dnn} have been shown to struggle with out-of-distribution generalization~\cite{damour2020underspecification, endocv2021_review, shortcut_learning, cleverhans, challenges_deploying}. This can be attributed to the fact that the features \glspl{dnn} learn are not necessarily causally related to the problem they intend to solve~\cite{causality}, but rather simply predictive with respect to the training data. The simplest manifestation of this behaviour is \textit{overfitting}, where the \gls{dnn} essentially memorizes the training data. While overfitting can be managed effectively through for instance regularization~\cite{l2_reg, batchnorm} and data augmentation~\cite{generalization_datamod}, this does not represent a complete solution. While these methods do improve generalization, they contribute primarily to increasing \glsfirst{ind} accuracy. Well regularized training pipelines have in several case studies been shown to fail to generalize. This can be attributed to \textit{shortcut learning}~\cite{shortcut_learning}, where the model learns features that are highly predictive for both the \gls{ind} training, validation, and test sets, but which result in poor performance in \gls{ood} contexts. This has for instance been exemplified in a study on \gls{dnn}-based pneumonia detection, wherein the model learned to associate pneumonia with hospital-specific watermarks rather than clinically relevant markers~\cite{pneumonia}. In another study, it was shown that a \gls{dnn} trained to detect  skin cancer and other dermatological conditions was basing its predictions mainly on  skin cancer skin markings rather than diagnostically relevant features~\cite{skin_shortcut}. This behaviour can be understood as a direct consequence of \textit{underspecification}~\cite{damour2020underspecification}, i.e. that there exists a large space of models that any given training algorithm can return for a given finite dataset, each characterized by different learned of sets predictive patterns observable in the training dataset, but nevertheless similar levels of accuracy on \gls{ind} data. 

In practice, this often manifests as poor accuracy on data that exhibits differing characteristics than the training data, i.e. \gls{ood} data~\cite{unifyingdatashift}. When such data is encountered, it is referred to as a \textit{distributional shift}, defined formally as follows:
\begin{equation}
    P_{train}(x,y) \neq P_{test}(x,y)
\end{equation}
A given datum \(x\) can be considered \gls{ood} if it lacks sufficient support in the training data, i.e. \(P_{train}(x)\approx 0\). Data that is drawn under shift can therefore be considered \gls{ood}. 

\subsection{Verification of Deep Learning Systems}
The sensitivity of \glspl{dnn} to distributional shift has been relatively well understood since their inception~\cite{unifyingdatashift, deep_learning_book}. The failure of \glspl{dls} in deployment conditions is therefore not necessarily only attributable to this sensitivity, but also a lack of robust praxis for \gls{dnn} evaluation. Existing approaches do not sufficiently account for the sensitivity of \glspl{dnn} to distributional shift, and instead typically involves simply collecting metrics from an unseen partition of the training dataset~\cite{large-scale-ml-deployment}. This typically results in test datasets by a high degree of similarity to the training set, especially as they often are created through random splits~\cite{testing_ml, tessting_ml_2}. However, there has been some work towards addressing these shortcomings, in particular with regards to the curation of informative test sets. DeepXplore~\cite{deepxplore}, for instance, involves synthetically generating test cases based on \gls{ood} seed inputs, and has been shown to increase testing coverage. It has been contested whether or not generative approaches in fact yield test cases representative of deployment scenarios or if they simply yield increased support and diversity within the training distribution~\cite{testing_ml}. 

\subsection{Distributional Shift Detection and Runtime Verification}
As the incidence of distributional shift can incur significant consequences during deployment, a growing body of work has emerged on the detection of distributional shift~\cite{dsd_survey}. These approaches are typically intended as a form of \gls{rv}. When a \gls{ood} detector detects that a shift has occurred, one may activate certain fail-safes or fallback measures in order to mitigate the consequences of an incorrect prediction. Generally, these approaches involve characterizing the distribution \(P_{train}(x,y)\) and the probability \(P_{train}(x_{test})\), often through the computation of surrogate features for the divergence \(P_{train}(x,y)||P_{test}(x,y)\). The first approaches to this end involved comparisons of maximum softmax probabilities~\cite{dsd_baseline}. More sophisticated approaches include energy-based approaches~\cite{dsd_energy}, approaches based on generative modelling~\cite{typicality_dsd}, gradients~\cite{gradients_distshift} or distance-based approaches~\cite{knn_dsd, mahalanobis_dsd}, with each reporting relatively high accuracy. While these methods can effectively detect the incidence of distributional shift and mitigate its effects to a certain extent, there are currently no publicly known methods which explicitly leverage their potential in the context of evaluation. 

\section{Probabilistic Evaluation and Risk Assessment}\label{ch:framework}
In this section, we outline the theory and implementation behind our proposed method. Our approach is heavily inspired by \gls{pra}, a methodology often used in other high stakes engineering disciplines such as aerospace engineering~\cite{pra_aviation} and nuclear safety engineering~\cite{pra_nuclear}. Central to our approach and \gls{pra} in general is the concept of an \textit{event tree}, which represents the probability of an event occurring and the conditional probabilities of its downstream effects. Whereas in other engineering disciplines these events are typically measurable and well defined (e.g. mechanical/electrical failures), the events in \gls{dls} domains are uniquely determined by the data and its nature, which is more difficult to characterize. Our principal contribution lies in the observation that data events can be characterized to a certain degree of accuracy through a corresponding detector, and that such detectors can be utilized towards computing the probabilities in the event tree. By virtue of their construction, it is possible to traverse the event tree in order to compute accuracy estimates given the estimated branch probabilities. Additionally, each leaf node may be associated with a weight in accordance with the cost of the corresponding outcome. For a medical system, for instance, these weights may be informed by figures taken from clinical economics~\cite{clinical_economics}, and correspond to measures of cost-to-life. In the following, we will outline the construction of event trees and discuss how their constituent probabilities may be computed. 

\subsection{Building Event Trees}
    The event trees we construct as part of our method are effectively a means of structuring evaluations according to the properties of the inputs. Each node in the tree is characterized by transition probabilities for its child nodes, and optionally an associated cost for risk assessment. The root node of these trees corresponds to the probability that the input is characterized by a given property (in our case - distributional shift). As we will discuss further in \Cref{se:pe}, the event trees presuppose the implementation of an event detector \(d_e(x)\) --- e.g. a \gls{ood} detector --- the verdicts of which are used to estimate these root probabilities. We outline two prototype event trees: the first simply considers the probability of correct predictions given the incidence of a shift, and the second additionally accounts for the verdict of event detector if they are for example used to activate fail-safes. Each node in the trees can also optionally be associated with a set cost, such that the traversal instead yields a risk estimate, which we explore in the context of Polyp segmentation in \Cref{se:case_study}
    
    In this paper, we primarily study the distributional shift event and its effect on visual systems, though we highlight that a risk tree may be constructed for any event as long as it is possible to implement a corresponding detector. Other examples of detectable events include adversarial attacks~\cite{adversarial_fourier, adversarial_subnetwork, adversarial_intrinsic} and \gls{llm} prompt violations~\cite{llm_monitoring_i1}.  It should also be noted that both of these trees are intended as generic bases, and that further branching may be appropriate depending on the domain. We discuss these matters further in \Cref{ch:discussion}. 
    
    \subsubsection{Base Event Tree}
\begin{figure}
        \centering
        \includegraphics[width=0.5\linewidth]{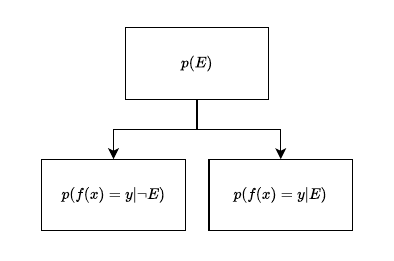}
        \caption{Base Event tree for a \gls{dnn}.}
        \label{fig:simple_et}
    \end{figure}

    Fundamentally, we can consider a \gls{dls} to be a function \(f\) that maps input data \(x\) with class \(y\) to a prediction \(\hat{y}\), i.e. \(f:x\rightarrow \hat{y}\). We can thus express the likelihood of a positive outcome across the space of input data \(X\) --- i.e. a correct prediction --- as \(p(\hat{y}=y|x\in X)\). With conventional validation, this likelihood is implicitly estimated for a test set, is thus conditioned on a subspace of \(X\), namely \(X_{test} \subset X_{ind}\). Since \(X = X_{ind} \cup X_{ood}\), the resulting estimates are flawed. This probability can, however, be decomposed into terms of the conditional probabilities for the \gls{ind} and \gls{ood} spaces:
    \begin{align}\label{eq:risk}
        p(\hat{y}=y|x\in X) &= p(x \in X_{ind})p(\hat{y}=y|x\in X_{ind}) \nonumber\\&+ p(x \in X_{ood})p(\hat{y}=y|x\in X_{ood})
    \end{align}
    This corresponds to a traversal of an event tree of height 2, shown in \Cref{fig:simple_et}.

    \subsubsection{Event Tree with Shift Detectors} 
\begin{figure*}
        \centering
        \includegraphics[width=0.7\linewidth]{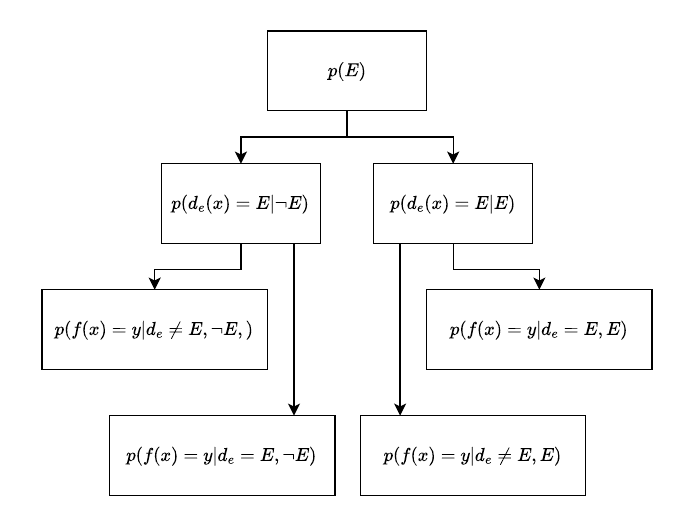}
        \caption{Event tree where \(d_e\) verdicts are utilized as RV.}
        \label{fig:event_tree_dsd}
    \end{figure*}

    Since our methods require a suitable event detector, it is logical to assume that the event detector verdicts may be used to modify system state, such as by activating failsafes or flagging the inputs for manual intervention. This effectively adds another layer to the tree -- corresponding to whether or not a given shift is detected by the \gls{ood} detector -- along with the associated costs involved in successfully detecting these shifts and activating a failsafe. This introduces two new outcomes: the \gls{ood} detector may correctly predict a distributional shift and activate a failsafe, or it may incorrectly predict a distributional shift and activate a failsafe where none was necessary. This also modifies the leaf probabilities, as each probability is now conditioned not only on the incidence of the event, but also whether or not this shift is successfully detected. This is shown in \Cref{fig:event_tree_dsd}. 
    
\subsection{Likelihood Estimators}\label{se:prob_est}
Estimating the overall risk requires multiplying the probability of each leaf node in the event tree with its associated cost. These leaf probabilities are in turn the product of the antecedent probabilities. Each of these probabilities therefore has to be estimated in a robust manner. In general, we distinguish between three types of probabilities:
\begin{enumerate}
    \item \textbf{Event probabilities} \(p(E)\): which denotes the probability of an external event happening, for instance encountering \gls{ood} data, adversarially generated data, or otherwise external changes to the execution environment that system support components such as \gls{ood} detectors and other forms of \gls{rv} may detect. These probabilities can be estimated via the analysis of \gls{rv} monitor traces, adjusted to account for verdict uncertainty. 
    \item \textbf{Conditional Detection Probabilities} \(P(d_{e}(x)=E | \{C\})\), which denote the probability that a detector \(d_c\), for instance a \gls{ood} detector or other \gls{rv} algorithm, correctly detects the incidence of these events. These probabilities can be thought of as accuracy metrics for the \gls{rv} algorithms, as well as the accuracy estimates for the \glspl{dnn} itself conditioned on the data for a given event chain. These probabilities can be computed by computing accuracy metrics for the given \gls{rv} component, e.g. \gls{tpr} and \gls{tnr} for \gls{ood} detectors. 
    \item \textbf{Conditional Accuracy} \(P(f(x)=y | \{C\})\), i.e. the probability of the \gls{dnn} yielding correct predictions conditioned on the antecedent events and detection outcomes. These probabilties can be estimated by computing accuracy metrics on specific partitions of the data. 
\end{enumerate}

\subsubsection{p(E)}\label{se:pe}

There is an emerging body of research implementing \gls{rv} algorithms for use in \glspl{dls}~\cite{enase_rv}. These algorithms are intended to detect the incidence of events that might negatively affect the system, and are typically implemented as auxiliary classifiers characterized by some \gls{tpr} and \gls{tnr}. By analysing traces of verdicts from these detectors, we can estimate the probability of the events they are intended to detect. This can be achieved by computing the frequency of positive verdicts, accounting for the uncertainty in the detector outputs. 

For a trace of detector verdicts \(t_{d} = [d_{e}(x_1), d_{e}(x_2),..., d_{e}(x_n)]\), 
the most straightforward estimate of the shift incidence is the empirical mean
\[
\bar{t} = \frac{1}{n}\sum_{i=1}^n d_e(x_i).
\]
However, since the detector is imperfect, this mean is biased with respect to the true underlying shift rate. Specifically, a true shift is detected with probability \(\mathrm{TPR}\), while a non-shift is misclassified as a shift with probability \(1-\mathrm{TNR}\). 
The relationship between the observed mean verdict \(\bar{t}\) and the true underlying shift rate \(p(E)\) is therefore
\[
\bar{t} = \mathrm{TPR} \cdot p(E) + (1 - \mathrm{TNR}) \cdot (1 - p(E)).
\]
Solving for \(p(E)\) yields the corrected estimate, known as the Rogan--Gladen estimator:
\[
\hat{p}(E) = \frac{\bar{t} - (1 - \mathrm{TNR})}{\mathrm{TPR} + \mathrm{TNR} - 1}.
\]

This corrected estimator directly adjusts the observed frequency of detector positives for systematic bias introduced by false positives and false negatives. Note that this assumes that the \gls{tnr} and \gls{tpr} are constant.
Unlike approaches that smooth or accumulate evidence across multiple traces, e.g. via Bayesian updating, here each trace window is corrected independently.  As a result, the estimator is more responsive to temporal changes in the underlying shift probability, at the cost of increased variance when the detector is weak (i.e., when \(\mathrm{TPR}+\mathrm{TNR}\) is close to 1).  In practice, this method is well-suited to dynamic environments where the probability of shifts is expected to change over time, and simplifies to the naive empirical mean when the detector operates at near-perfect accuracies (\(\mathrm{TPR}, \mathrm{TNR} \approx 1\)).

As it is natural and even expected that the environment will shift continuously, the length of the trace must be tuned such that is is appropriate for the domain in question. Lower trace lengths result in more uncertain estimates, but are more indicative of the current shift environment. Conversely, long trace lengths provide a more certain estimate for a long time horizon, and in effect marginalizes over the shift environment variability. In practice, this choice should be informed by the desired balance between short-term responsiveness and long-term risk forecasting. If, for instance, a given application requires that the system is interrupted immediately if it exceeds a specified risk tolerance, a low trace length should be chosen. If the risk assessment is instead intended to gauge the long-term benefit of deploying a given system, a long trace length may be more appropriate.

Beyond adjusting trace lengths, another feasible approach is to multiply the traces with a weighting function such that more recent verdicts are given more weight, for instance through an exponential decay function. We leave the exploration of more sophisticated weighting strategies as future work in the interest of brevity. 

\subsubsection{\(P(d_{e}(x)=E | E)\)}\label{se:peg}
Since it is unlikely that any given event detector exhibits perfect accuracy, it is necessary to account for its uncertainty during risk assessment. To this end, it is necessary to compute the conditional probability of a positive verdict being correct given the event occurring --- i.e. the detector \gls{tpr}--- and the probability of a negative verdict being correct given an event not occurring  --- i.e. the detector \gls{tnr}. These probabilities can be estimated as part of the initial evaluation of the corresponding detector. 
\subsubsection{\(P(f(x)=y | E)\) and \(P(f(x)=y | E, d_e(X)=E)\)}\label{se:pegg}
The accuracy of a neural network is largely dependent on the characteristics of the incoming data. Generalization failure can, for instance, largely be attributed to the prevalence of distributional shift. Network accuracy may equivalently be impeded by adversarial attacks or other events. It is as a result necessary to account for the probability of the network yielding correct predictions conditioned on the incidence of these events, and, if necessary, whether or not these events have been successfully detected. 

Consequently, it is necessary to compute the conditional accuracies of the \gls{dnn}. We achieve this by simply computing accuracy metrics on partitions of the detector validation data extracted according to the pertinent conditions. For the event tree in \Cref{fig:simple_et}, this entails computing the accuracy for an \gls{ind} validation set and an \gls{ood} calibration dataset. Lacking the latter, it is also possible to create such a dataset through synthetic augmentation, e.g. by applying degrees of additive noise. For the event tree in \Cref{fig:event_tree_dsd}, it is also necessary to account for the shift detector verdict, i.e. the \gls{tpr} and \gls{tnr} of the \gls{ood} detector given that the data is \gls{ind} or \gls{ood}. If a set of conditions does not yield data, the probability of it occurring is assumed zero. 

\subsection{Applying the Event Tree}
It is worth emphasising again that our approach is intended to be used \textit{at runtime}, since \(p(E)\) may evolve dynamically over the course of deployment. Applying the event tree thus involves iteratively updating this estimate. This can be achieved by implementing the verdict traces as a queue. When new data is encountered, the oldest verdict is discarded in place of the current verdict. The estimate for \(P(E)\) is then updated accordingly. 

This does, however, mean that the estimate of \(p(E)\) will likely be characterized by some degree of error until a complete trace is available. The application of our methods is thus ideally coupled with a sort of pilot study, both to ensure the precision of the accuracy estimates and in order to select a sufficient prior estimate for \(p(E)\). 

\section{Methodology}\label{ch:methodology}
In this section, we outline the methodology used to assess the efficacy of \gls{name}. 

\subsection{Metrics and Evaluation}
To assess the errors in the estimation of both the event rate, expected accuracy, and expected risk, we utilize \gls{mae}, defined as:
\begin{align*}
    MAE & = \frac{1}{N}\sum_i^N |y_i-\hat{y}_i|\\
\end{align*}

For the accuracy estimation experiments, we implement a baseline that yields the estimated \gls{ind} validation accuracy for all data. While taking an average between this accuracy and the accuracy on an available \gls{ood} test set can be argued is a more solid baseline, this implicitly assumes a shift incidence rate of 0.5, and thus simply represents a shift with respect to the rate-wise error rather than constituting a superior estimate overall. 

\subsection{Benchmarking Datasets}
In order to effectively validate our approach, we require datasets which contain partitions characterized by distributional shift. To this end, we utilize four classification datasets primarily intended for domain adaptation benchmarking, shown in \Cref{tab:benchmarking_datasets}. Each dataset is composed of different contexts/data environments, each of which can be considered \gls{ood} to one another. For the Office31, OfficeHome, and NICO++ datasets, we extract training data, \gls{ind} validation data, and \gls{ind} test data from one context, with the remaining data being used as \gls{ood} data. As our proposed event trees assume that \gls{ood} data is characterized by approximately the same probability of correct prediction, we compute metrics for each fold independently such that any such discrepancies are made evident as part of our analyses. For all of our experiments, we keep the same \gls{ind} context for each dataset, chosen at random. While it can be argued that training separate models for each context and cross-validating represents a more robust methodological approach, we reiterate that our analyses are not concerned with the precise values of network accuracy inasmuch as the errors in their estimation. 

\begin{table*}[htb]
    \centering
    \caption{Overview of benchmarking datasets.}
    \input{tables/benchmarking_datasets} 
    \label{tab:benchmarking_datasets}
\end{table*}

\subsection{OOD detectors}\label{se:ed}
In this work, we primarily study \gls{ood} detectors, which can be interpreted as an event detector intended to detect the incidence of distributional shift. To sufficiently evaluate our methods we thus select the following implementations of \gls{ood} detectors:
\begin{table}[!htb]
    \centering
    \input{tables/ood_detectors}
    \caption{\gls{ood} detector implementations.}
    \label{tab:ood_detectors}
\end{table}
This selection was informed by the taxonomy outlined in~\cite{dsd_survey} and is intended to reflect the general scope of approaches utilized in the field. We acknowledge that there exist implementations that outperform the above, but have foregone implementing these in the interest of simplicity of analysis. Where applicable, we show results for the best performing implementation for each dataset among those outlined above. 

The balanced accuracies attained by these \gls{ood} detectors for each of the datasets is shown in \Cref{tab:ood_detector_accuracy}. Note that we utilize the most accurate \gls{ood} detector for each dataset in each of our analyses.
\begin{table}[!htb]
    \centering
    \input{tables/ood_detector_accuracy}
    \caption{Balanced accuracies of \gls{ood} detectors on each dataset.}
    \label{tab:ood_detector_accuracy}
\end{table}

Since the efficacy of our method is dependent on detector accuracy, we also perform sensitivity analyses with respect to a synthetic detector with a given \gls{tpr} and \gls{tnr}. The detector is implemented such that it yields a correct positive prediction with a uniform probability of \(p=TPR\) and likewise a correct negative prediction with a uniform probability of \(p=TNR\). We collect data across each combination of \gls{tpr} and \gls{tnr} in the set [0.0, 0.1, 0.2,0.3,0.4,0.5,0.6,0.7,0.8,0.9,1.0], such that the balanced accuracy exceeds 0.5, i.e. \(\frac{1}{2}(tpr+tnr)>0.5\).

\subsection{Model Training}
We train a ResNet101~\cite{resnet} for each of the benchmarking datasets and a DeepLabV3+~\cite{deeplab} for the polyp segmentation case study. The training configuration we used is summarized in \Cref{tab:training}. While other architectures may yield improved accuracy, the focus of this paper is not on maximizing accuracy, but rather on computing credible accuracy estimates. 

\begin{table}[htb]
    \centering
    \caption{Training Configuration.}
    \input{tables/training_configuration}
    \label{tab:training}
\end{table}

To illustrate both the extent of generalization failure in our chosen datasets, and in order to contextualize our results in \Cref{ch:results}, we provide accuracy estimates for each dataset and fold in \Cref{tab:dataset_results}. Note that for the polyp segmentation datases, we define an accurate prediction as a prediction with an IoU score exceeding 0.5 for individual samples. For batched experiments, we consider a correctly predicted batch one for which the batch-wise accuracy exceeds the 1st percentile observed batch-wise \gls{ind} accuracy. 
\begin{table}[h]
    \centering
    \input{tables/classifier_accuracy}
    \caption{Performance metrics for different datasets and shifts.}
    \label{tab:dataset_results}
\end{table}
\subsection{Deployment Simulation}
To simulate deployment conditions, we construct data streams composed of batches of a given size and collect metrics from a range of batch sizes, i.e. (1, 8, 16, 32, 64). With a batch size of 1, this simply entails analysing each input sample individually and can be considered the default operating mode. However, since practical implementations of deep learning systems typically operate on batches for the sake of computational efficiency, it can also be considered natural that further analysis is performed on a batch-wise basis. Many distributional shift detection algorithms also operate in this mode~\cite{typicality_dsd, failing_loudly}. We thus assume that the distributional shifts also occur on a batch-wise basis. For simplicity of analysis we consequently assume that each batch consists of exclusively \gls{ind} or \gls{ood} samples. Since shifts of practical consequence occur in clusters, we contend that this assumption to a large extent holds across most deployment scenarios. We consider a batch to be correct if the mean accuracy for a given batch is greater than or equal to the 1st percentile mean batch-wise accuracy in the validation data. 

For a sequence of batches, we assume that the events are Bernoulli distributed, with a set Bernoulli expectation. While it is not given that distributional shifts on batches are in fact Bernoulli distributed, we reiterate that our approach is concerned with estimating the probability of the network yielding incorrect predictions for a given time horizon rather than for individual samples, which constitutes an estimate of the overall \textit{frequency} of events over said time horizon. The distribution of the events is as a result of minor consequence. When it is specified that an event has occurred, we replace the corresponding batch in the stream with a batch of \gls{ood} samples. 

\section{Results}\label{ch:results}
We validate the efficacy of our approach through several experiments. We first assess the errors associated with parameter estimation. Then, we consider the extent to which the accuracy over a given trace can be accurately predicted using our methods. To illustrate the significance of our work, we also provide examples of insights that may be inferred using our framework in practical scenarios. 

\subsection{Event Tree Parameter Estimation}
Our approach is based on the assumption that we can compute accurate probability estimates with respect to each branch of the given event trees. The error of any measures extracted from the event tree --- therein accuracy and risk estimates --- are dependent on the error of these probabilities. We thus assess the error rates of each of the probability estimation methods outlined in \Cref{ch:framework}. 

\subsubsection{Rate Estimation}
A significant advantage of our framework over conventional verification is that it can account for varying incidence rates of distributional shift at runtime. The accuracy of the shift frequency estimates is as a result of significant importance for an accurate risk assessment. We thus assess the accuracy of these estimates through sensitivity analyses with respect to \gls{ood} detector performance. As outlined previously, we utilize synthetic shift detectors in order to fill the input space and contextualize the results in terms of several popular \gls{ood} detector algorithms.

\begin{figure}
    \centering
    \includegraphics[width=0.50\linewidth]{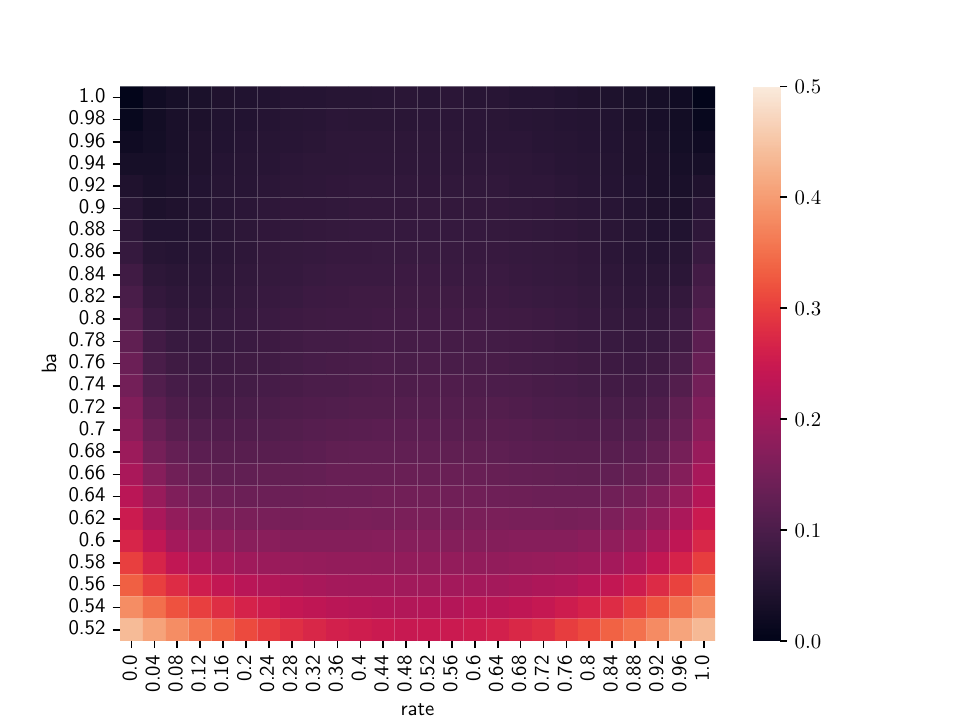}
    \caption{Average Rate estimation error across ground-truth rates as a function of detector BA.}
    \label{fig:est_sensitivity}
\end{figure}

We first assess the rate estimator on Bernoulli distributed data across a span of ground-truth rates, the results of which are shown in \Cref{fig:est_sensitivity}. Overall, the estimation methods seems to perform reasonably well given sufficiently high \gls{ood} detectors accuracy, with detectors of accuracies higher than \(\sim0.75\) typically not exceeding errors of 0.1.  

\begin{figure*}[!htb]
    \centering
    \includegraphics[width=0.7\linewidth]{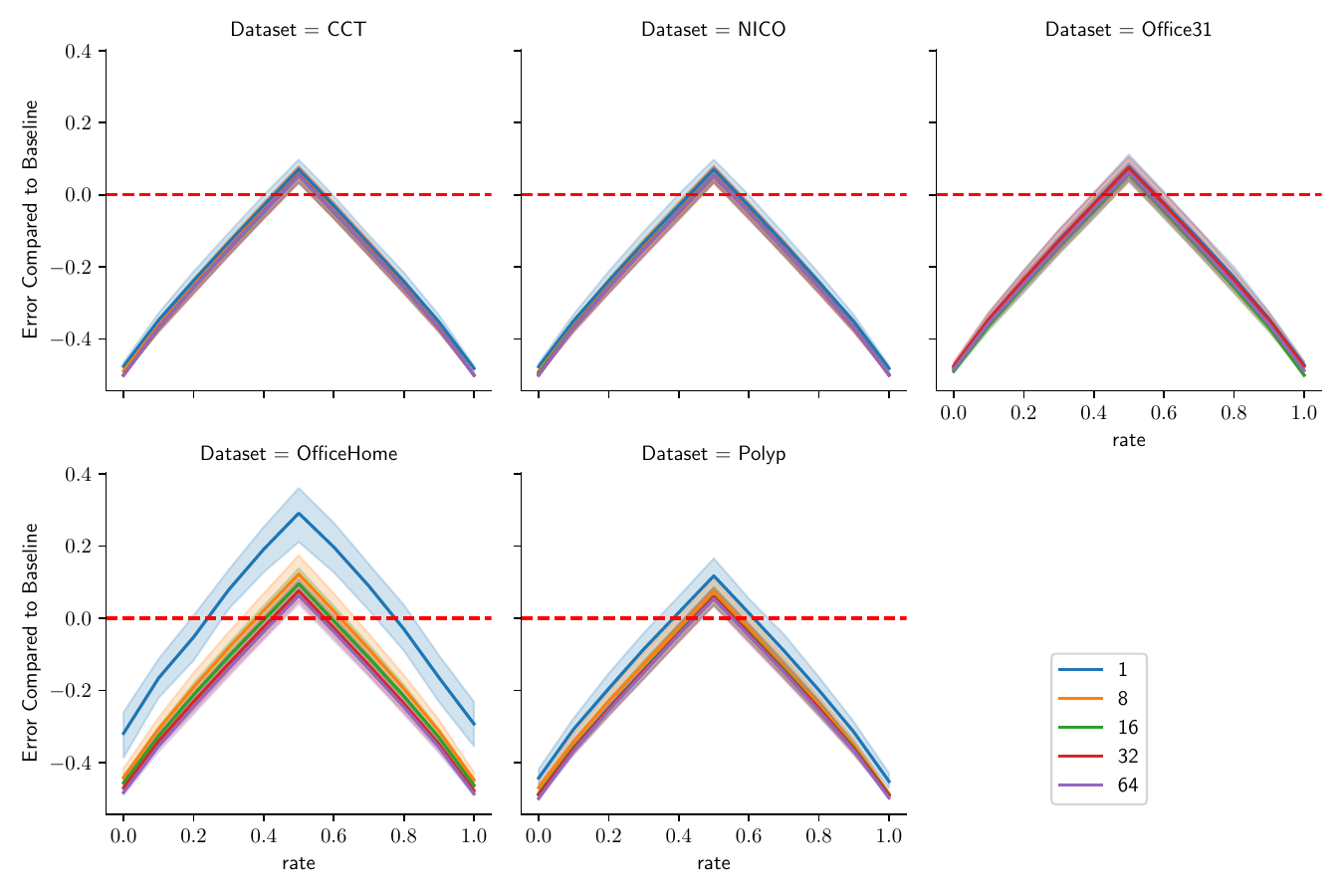}
    \caption{Rate estimation errors for the best \gls{ood} detector implementations across batch sizes. }
    \label{fig:dsd_rate_errors}
\end{figure*}

To contextualize this further, we provide error plots for a selection of popular \gls{ood} detectors algorithms in \Cref{fig:dsd_rate_errors} for a batch size of 32. In general, there is at least one \gls{ood} detector which attains reasonably low error rates (<0.1) for each dataset. The curvature of the relationship can be attributed to the balance between the \gls{tpr} and \gls{tnr} for the detectors. 

\subsubsection{Prediction Accuracy Estimation}
\begin{figure*}[h!tb]
    \centering
    \includegraphics[width=0.7\linewidth]{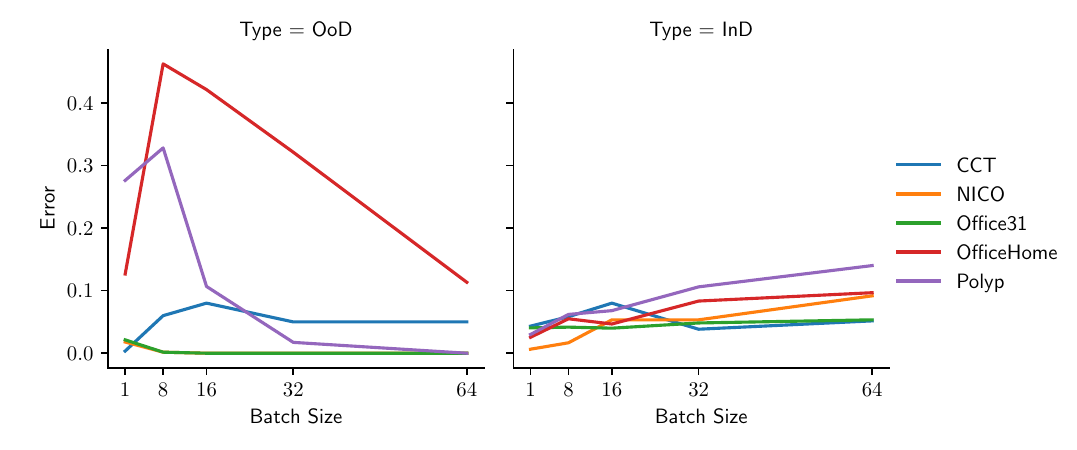}
    \caption{Accuracy estimate errors across batch sizes for each dataset. The extent of the error bands represent the minimum and maximum errors across the pertinent dataset folds. }
    \label{fig:classifier_errors}
\end{figure*}

The efficacy of our approach is predicated on accurate estimates of network accuracy, given the \gls{ood}-ness of the data. As a result, it is necessary to quantify the error in these estimates. As discussed in \Cref{ch:methodology}, our method assumes a given batch size.  We thus provide cross-validated error estimates across a span of batch sizes in \Cref{fig:classifier_errors} for the event tree in \Cref{fig:simple_et}. For all datasets, there is little discrepancy between the \gls{ind} validation and test data. For the NICO++, Office31, and CCT datasets, we observe that there is little discrepancy between the accuracy estimates across the different folds of \gls{ood} data as well. For the Polyp and OfficeHome datasets, there is a greater degree of variability, though this is largely attributable to the relatively higher accuracy on the Product context in the latter and the relatively lower accuracy on the Etis-Larib dataset in the former.  

\begin{figure*}[h!tb]
    \centering
    \includegraphics[width=0.7\linewidth]{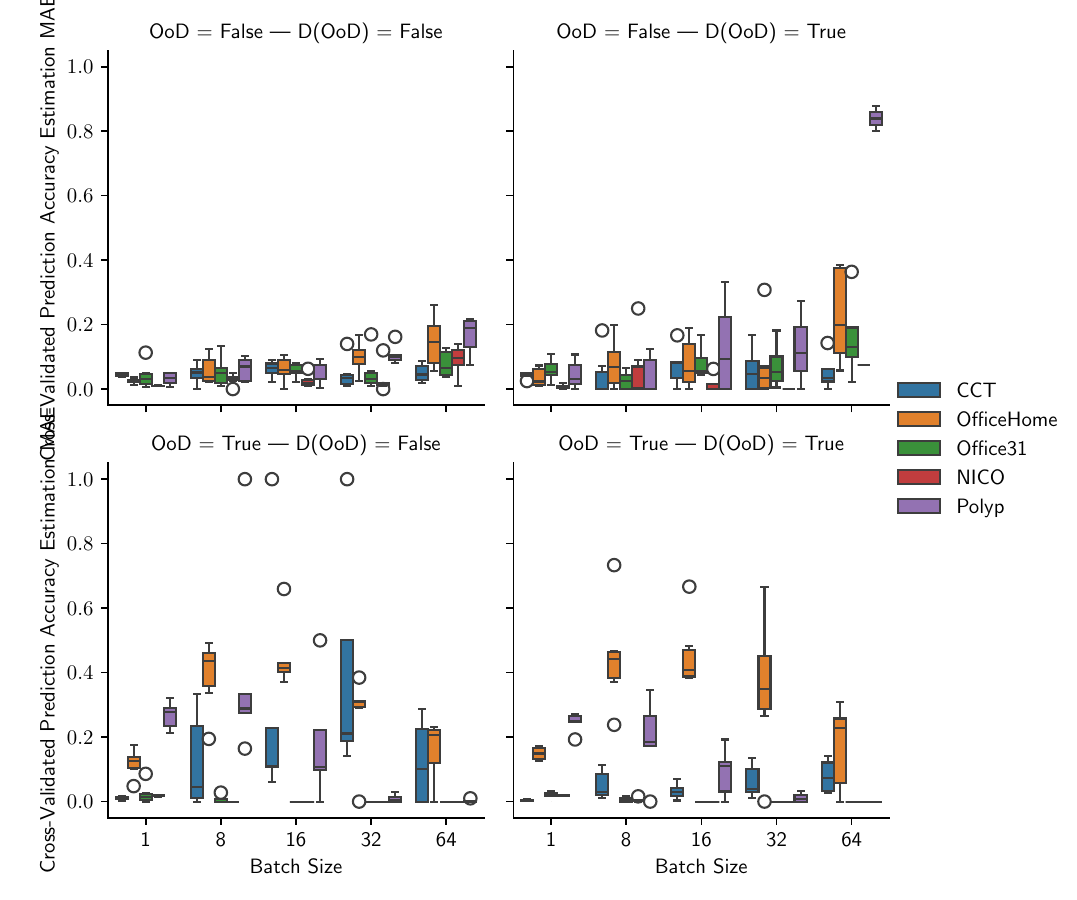}
    \caption{Accuracy estimation errors conditioned on the \gls{ood}-ness of the data and whether or not it is detected correctly, i.e. the leaf node probabilities in \Cref{fig:event_tree_dsd}.}
    \label{fig:errors_dsd_tree}
\end{figure*}

For the  event tree in \Cref{fig:event_tree_dsd}, the analysis is somewhat more involved. For one, this event tree is constructed such that it account for discrepancies in prediction accuracy conditioned on both the incidence of a shift and whether or not this shift has been detected. It may or may not be the case that prediction accuracy is independent of the monitor verdicts given a shift has occurred, i.e. \(f(x)=y\ind d_e(x)=E|E\). The errors in these estimates thus depend on degree of dependence, which is a property of whichever event detection algorithm is utilized. As a result, we assess these errors with respect to the implementations outlined in \Cref{se:ed}. The results are shown in \Cref{fig:errors_dsd_tree}. For \gls{ind} data that is correctly detected as such, the errors are generally quite low. Misclassified \gls{ind} data are associated with slightly lower errors. \gls{ood} data are, similarly to the base event tree, associated with lowe errors for the Office31, NICO++, and CCT datasets, and high for the Polyp and OfficeHome datasets. 

\subsubsection{Event Detector Accuracy Estimation}
\begin{figure*}[!htb]
    \centering
    \includegraphics[width=0.7\linewidth]{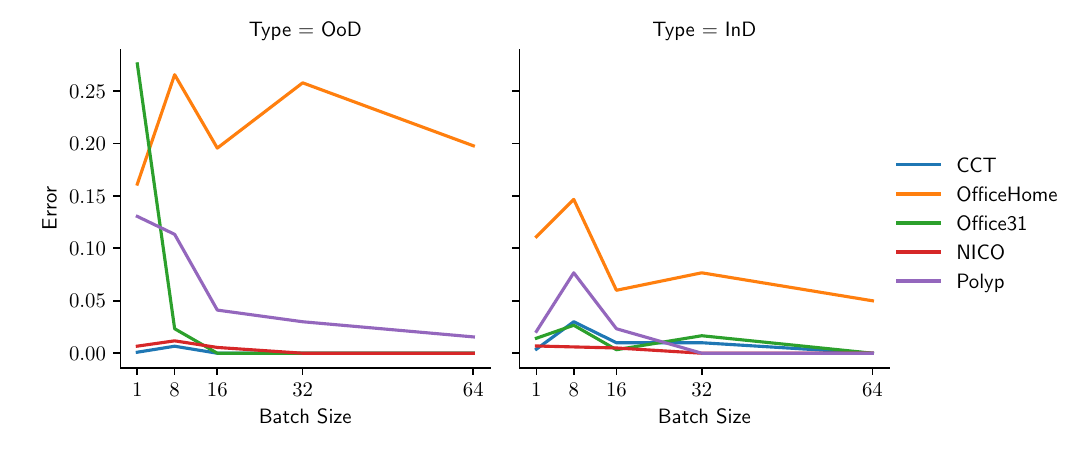}
    \caption{Errors associated with \gls{ood} detector \gls{tpr} and \gls{tnr} estimates. }
    \label{fig:dsd_acc}
\end{figure*}

We assess the \gls{ood} accuracy parameter estimation by computing cross-validated errors with respect to the detector accuracy across the \gls{ood} folds used to calibrate the \gls{ood} detectors, for the best \gls{ood} detector implementations for each dataset.  The results are shown in \Cref{fig:dsd_acc}. Overall, the estimation errors are fairly low for \gls{ind} data (i.e. \gls{tnr} and \gls{fnr}). For \gls{ood} data, only NICO and CCT are associated with consistently low error, with Office31 being characterized by large errors for a batch size of 1, OfficeHome consistently being associated with high errors across batch sizes, and the polyp datasets being characterized by moderate error that is reduced with increasing batch size.

\subsubsection{Summary}
Thus far, we have assessed the error rates for each of the parameter estimators in our two event trees. Overall, we have shown that:
\begin{itemize}
    \item Given a \gls{ood} detector of sufficient accuracy, the incidence rate of distributional shifts can be estimated with a high degree of precision. 
    \item If we stratify by \gls{ind} and \gls{ood} data, we can compute low error estimates of \gls{ind} network accuracy. \gls{ood} data, on the other hand, can be characterized by different levels of accuracy depending on the type and severity of the corresponding distributional shift, and thus any estimate will likely be characterized by a certain degree of error.
    \item The error of the \gls{ood} detector accuracy estimate depends on the degree to which its performance generalizes to different types of shift. In our experiments, the most accurate \gls{ood} detectors often exhibited a relatively high degree of generalization and thus lowest estimation error.
\end{itemize}
\subsection{Accuracy Prediction}
\begin{figure*}[!htb]
    \centering
    \includegraphics[width=\linewidth]{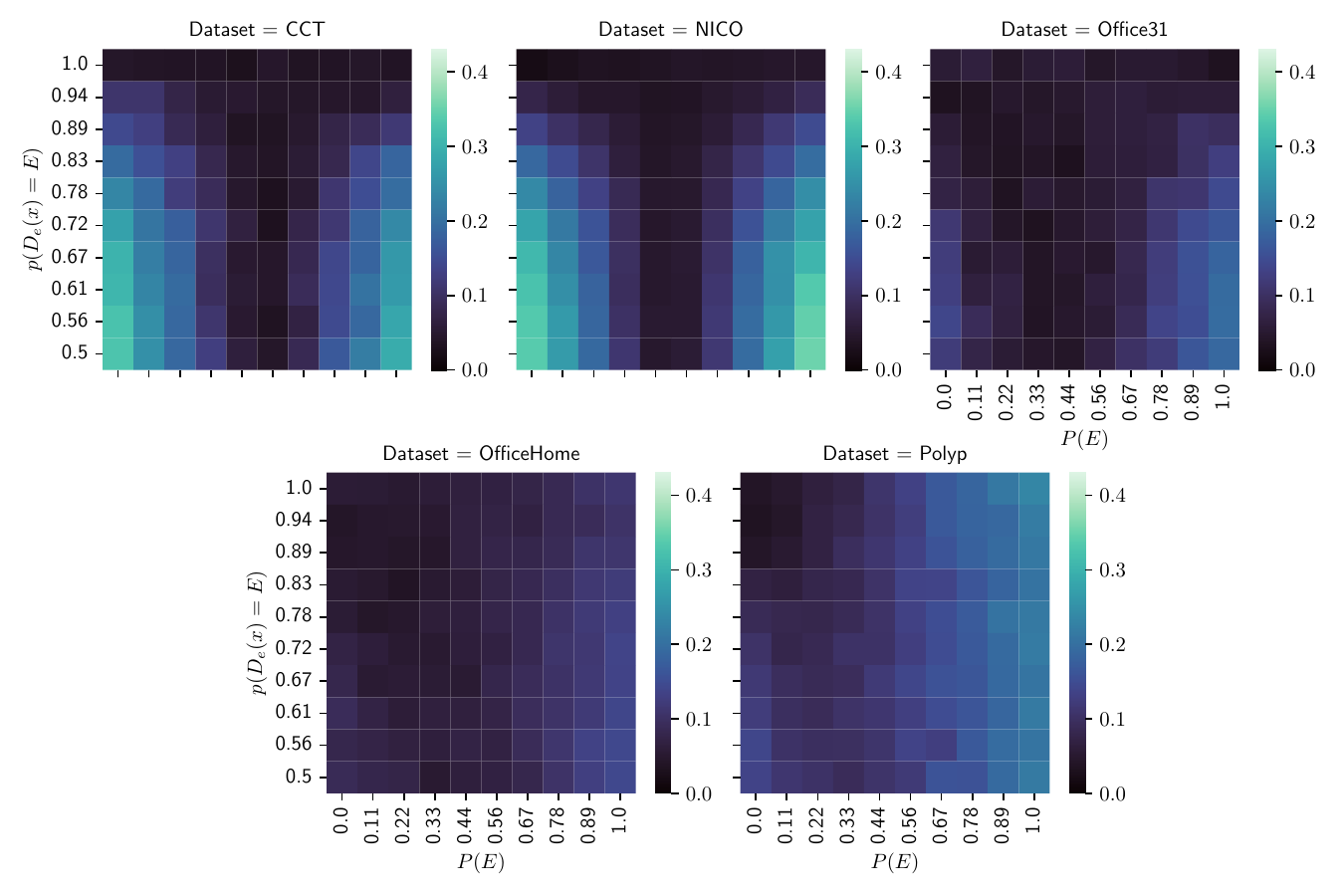}
    \caption{Cross-validated accuracy estimate errors as a function of rate and shift detector accuracy.}
    \label{fig:synth_dsd_results}
\end{figure*}

Thus far, we have assessed the errors in the the parameter estimates for our two event trees. In this section, we will assess the errors in the accuracy estimates that we extract from these event trees. To this end, we vary the incidence rates of distributional shift, and observe the difference in the predicted accuracy and the actual accuracy over a set time horizon. For the sake of generality, we first assess our methods with a synthetic \gls{ood} detector as defined in \Cref{se:ed}. The results are shown in
\Cref{fig:synth_dsd_results}, and indicate that our methods are capable of yielding highly credible accuracy estimates given a sufficient level of \gls{ood} detector accuracy. However, note that the resulting estimates are biased towards the observed accuracy in a given \gls{ood} calibration dataset, and thus that there is some error for the benchmarks where there are large discrepancies in the accuracy of different \gls{ood} folds, particularly for the OfficeHome dataset and Polyp datasets. We discuss this in more detail in \Cref{ch:discussion}.

\begin{figure*}
    \centering
    \includegraphics[width=\linewidth]{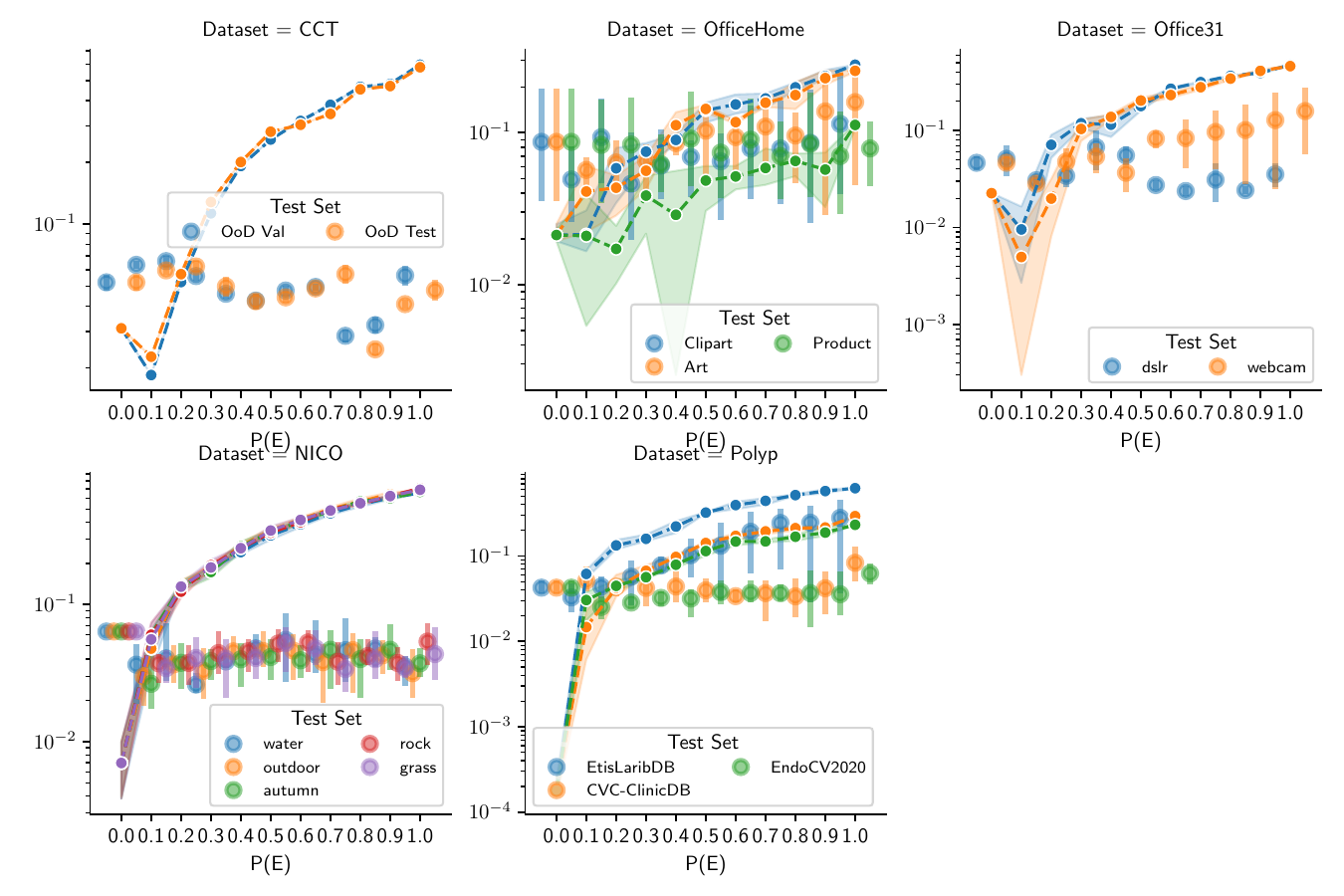}
    \caption{Cross-validated accuracy estimate errors for the tested \gls{ood} detectors as a function of rate, averaged over validation sets and batch sizes.}
    \label{fig:dsd_acc_errors_by_rate}
\end{figure*}

We further analyse our methods on the best performing \gls{ood} detector among the implementations outlined in \Cref{se:ed} for each dataset. We first consider the errors with respect to the event incidence rates. As mentioned in \Cref{ch:methodology}, we compare our results to a baseline which yields the estimated \gls{ind} validation accuracy for all data. The results are shown in \Cref{fig:dsd_acc_errors_by_rate}. Note that these results are averaged over batch sizes. A complete breakdown over errors across batch sizes can be found in \Cref{fig:bs_error_breakdown}. When the incidence rates of distributional shift are low, our estimates do not typically outperform the baseline. This is expected, however, as in these conditions the \gls{ind} validation data represents a veracious representation of the deployment conditions. With higher rates, however, our approach outperforms the baseline, often by an order of magnitude or more. There may be exceptions to this, however, notably if the \gls{dnn} encounters \gls{ood} data to which it succeeds in generalizing. This can be observed for the Product context in the OfficeHome dataset, which is associated with similar levels of accuracy as the \gls{ind} datasets as shown in \Cref{tab:dataset_results}. This results in higher errors since our approach assumes that all \gls{ood} data is characterized by similar levels of (low) accuracy. We discuss this further in \Cref{ch:discussion}. 

\begin{figure*}
    \centering
    \includegraphics[width=\linewidth]{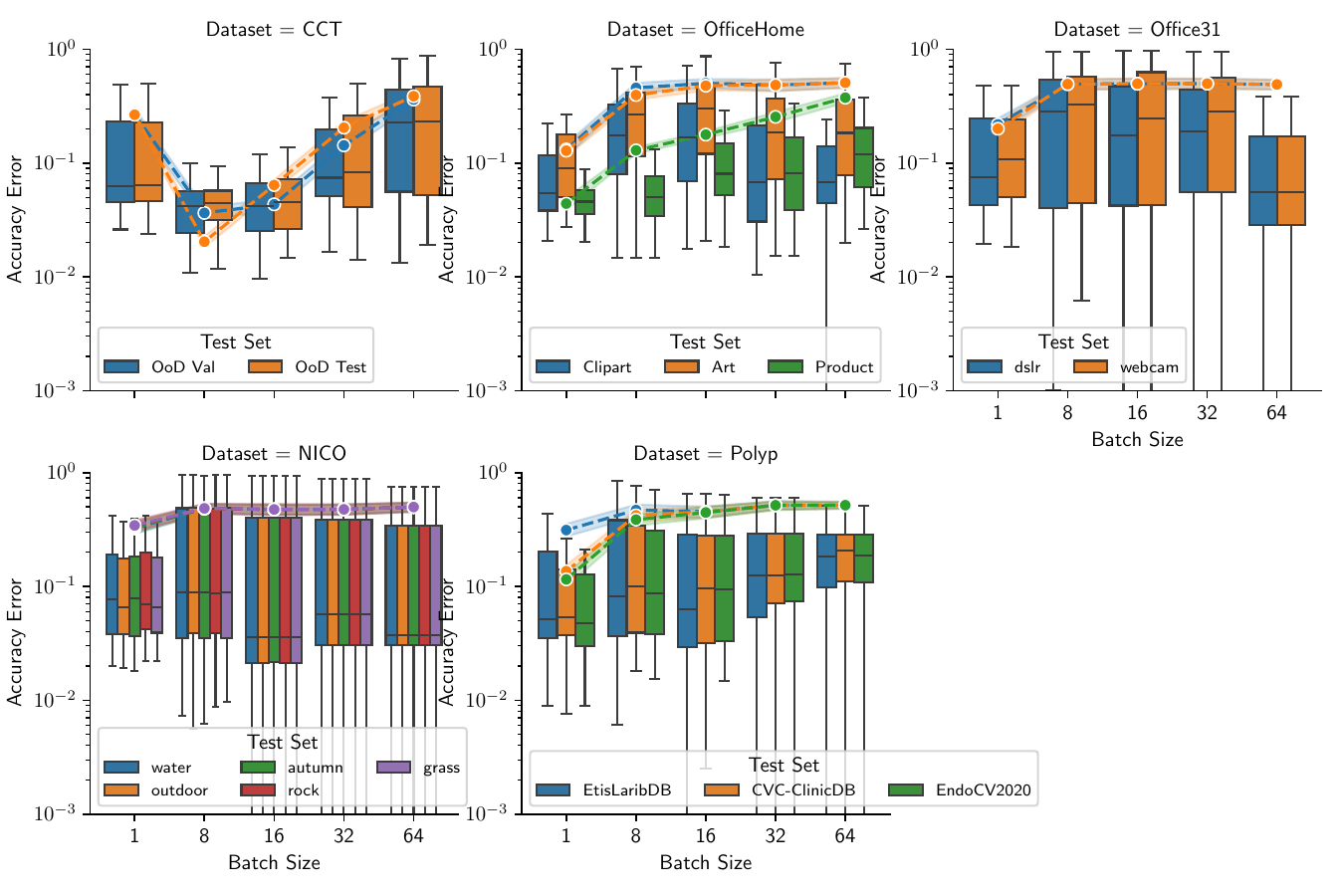}
    \caption{Accuracy estimate errors for the highest performing \gls{ood} detector implementations across batch sizes.}
    \label{fig:dsd_results}
\end{figure*}

As mentioned in \Cref{se:ed}, many implementations of \gls{ood} detectors are intended to be used in a batched configuration for optimal accuracy. We thus assess our approach across batch sizes in \Cref{fig:dsd_results}. Interestingly, while increasing batch size evidently increases \gls{ood} detector accuracy as shown in \Cref{fig:dsd_acc}, this does not necessarily always result in reduced accuracy estimation errors.
This apparent contradiction can be attributed to several factors: (i) larger batch sizes may improve shift detection but introduce greater variability in the conditional accuracy estimates due to the diversity of samples within each batch; (ii) the relationship between detection accuracy and overall accuracy estimation is non-linear, with diminishing returns after a certain detection performance threshold; and (iii) dataset-specific characteristics affect how batch composition influences error propagation through the event tree. This suggests that optimal batch size selection requires balancing shift detection performance against the precision of conditional accuracy estimates for each specific deployment scenario.

While the errors are indeed reduced for the NICO++ and CCT datasets, the errors remain relatively stable across batch sizes for the polyp datasets as well as for Office31, and shows an increasing trend for the OfficeHome dataset. This can be attributed to increased errors in the conditional prediction accuracy estimates for higher batch sizes.  

Note that, for these results, we assume uniformity in the batches. We include results for non-uniform batches in \Cref{fig:nonuniform} in \Cref{a:bu}. While our method also outperforms the baseline for non-uniform batches, the error margins are much larger in this case, to the extent that simply using a batch size of one results in lower errors. 

\subsection{Risk Assessment Case study: Polyp Segmentation}\label{se:case_study}
In \Cref{ch:results}, we have demonstrated empirically that our method permits the computation of accuracy estimates with a high degree of precision. Next, we will explore more concretely the benefits our approach can afford in the context of risk-estimation, using the Polyp segmentation benchmark as a case study. In particular, we demonstrate how our methods may be used to construct monitors of overall system specification attainment as expressed by runtime risk, as well as inform cost-benefit analyses of in development settings.

As mentioned in \Cref{ch:framework}, it is possible to associate each node in an event tree with an associated cost. For accuracy estimation, this simply entails setting the leaf node costs to 1 if the prediction is correct and zero otherwise. In many cases, however, it may be beneficial to distinguish between the costs of different outcomes. In this case study, we therefore associate each outcome with an economic cost, such that the system can be evaluated from a clinical-economics perspective. To this end, we provide an overview of each outcome and the associated cost in \Cref{tab:cost_table}. Note that while the table is largely informed by real-world clinical economics figures from the United States~\cite{polyp_lifetime_cost}, they are intended primarily as an illustrative example. All outcomes are associated with a cost of \$635, the cost of performing an endoscopy. A failed detection incurs a lifetime cost of \$6,100. If the \gls{ood} detector raises a positive verdict, it alerts the clinician that manual intervention is needed, for which we distinguish between necessary interventions --- where the model would predict incorrectly --- and unnecessary interventions, where the model would not predict correctly. We associate the latter with the base cost plus the probability and associated cost of a clinician failing to detect the polyp. We associate the former with the same cost, but subtract \$50 since the data is worth adding to the the training set and thus in effect constitutes an opporunity for improving the system.

\begin{table}[htb]
    \centering
    \begin{tabular}{cc}
    \toprule
         Outcome & Cost (\$) \\
        \midrule
         Correct Detection & 635 \\
         Necessary Intervention & 1905 \\
         Unneccessary Intervention & 1955 \\
         Failed detection & 6735\\
    \bottomrule
    \end{tabular}
    \caption{Breakdown of outcome-wise per-patient costs. }
    \label{tab:cost_table}
\end{table}

In \gls{pra}, it is common to define a level of \textit{maximum tolerable risk}. This is the point at which the risks associated with the system exceeds what is deemed acceptable. In this context, one may for instance assign threshold for maximum tolerable risk equal to the cost of simply always requiring manual intervention (i.e. \$1925). If the cost of a system implementing a \gls{dnn} exceeds this cost, one might as well after all not use it at all. 

\begin{figure}[htb]
    \centering
    \includegraphics[width=0.5\linewidth]{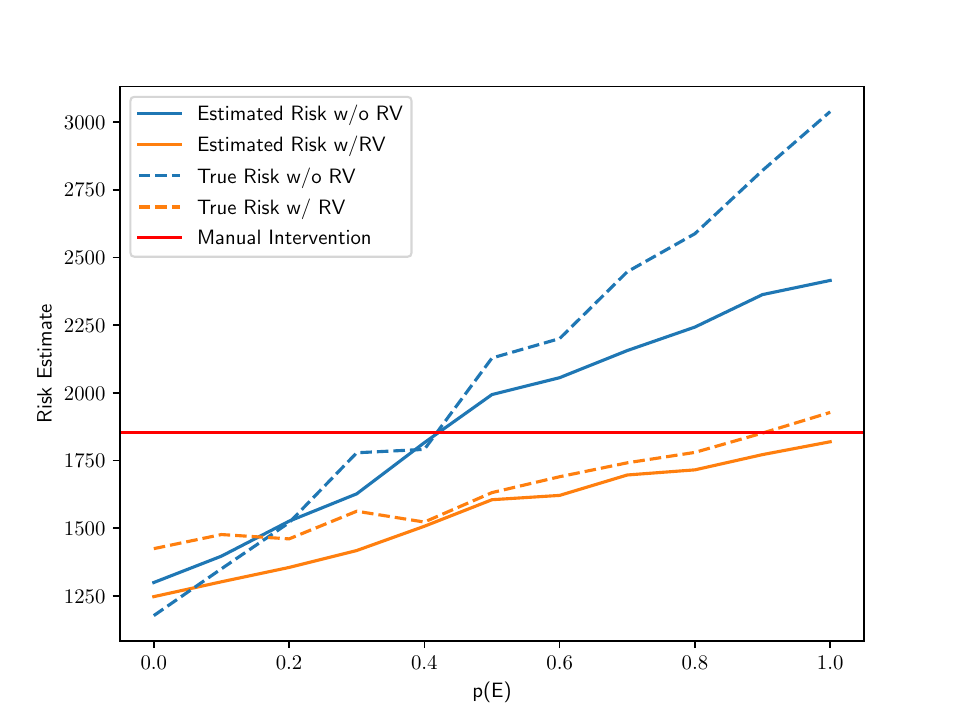}
    \caption{Estimated and actual risk vs incidence rate.}
    \label{fig:risk_estimation}
\end{figure}

To illustrate the utility our methods afford in this regard, we estimated risk as a function of the distributional shift incidence rate, and compare it to the actual costs and this maximum tolerable risk. We include plots for both event trees and corresponding system configurations. The results are presented in \Cref{fig:risk_estimation}, with the lines corresponding to the estimated risk and the extent of the areas corresponding to the true risk. 

First, it is worth noting that there is a clear distinction between the system which implements the detector verdicts as a form of \gls{rv} (i.e. calling for manual intervention given a positive verdict), and the system which does not. The former is for the most part associated with lower risk than the baseline, whereas the latter incurs greater risk beyond a shift rate of about 0.4. In effect, this means that a system that does not implement \gls{ood} detectors will, beyond this threshold, result in worse outcomes in the long term. For the system that implements \gls{ood} detectors, the costs only exceed manual intervention beyond a rate of around 0.9. 

The estimated risks differ somewhat from the true risks in this configuration. For the system implenting \gls{ood} detectors, thhe difference is relatively marginal. For the system which does not, however, the risk is significantly underestimated for high values if \(p(E)\). 

In addition to assessing the costs of the system, our methods can also be used to inform the development process. In industrial applications, the development of \gls{dls} is often characterized by ad-hoc practice~\cite{large-scale-ml-deployment}. Without sufficiently informative evaluation metrics, it is generally difficult to know whether a given modification to the model architecture, training algorithm, hyperparameter, etc. will add value to the system as a whole or not. A change which at first glance seems to improve network accuracy on a given test set may impede the accuracy of the system as a whole~\cite{backwards_compat}. In systems composed of multiple individual components --- e.g pipelined networks alongside event detectors --- it may also be difficult to determine how best to allocate development resources. Expressing the overall costs in terms of an event tree can assist in this regard, as it permits the calculation of the change in expected cost with respect to any change in its constituent probabilities. 

\begin{figure}
    \centering
    \includegraphics[width=0.5\linewidth]{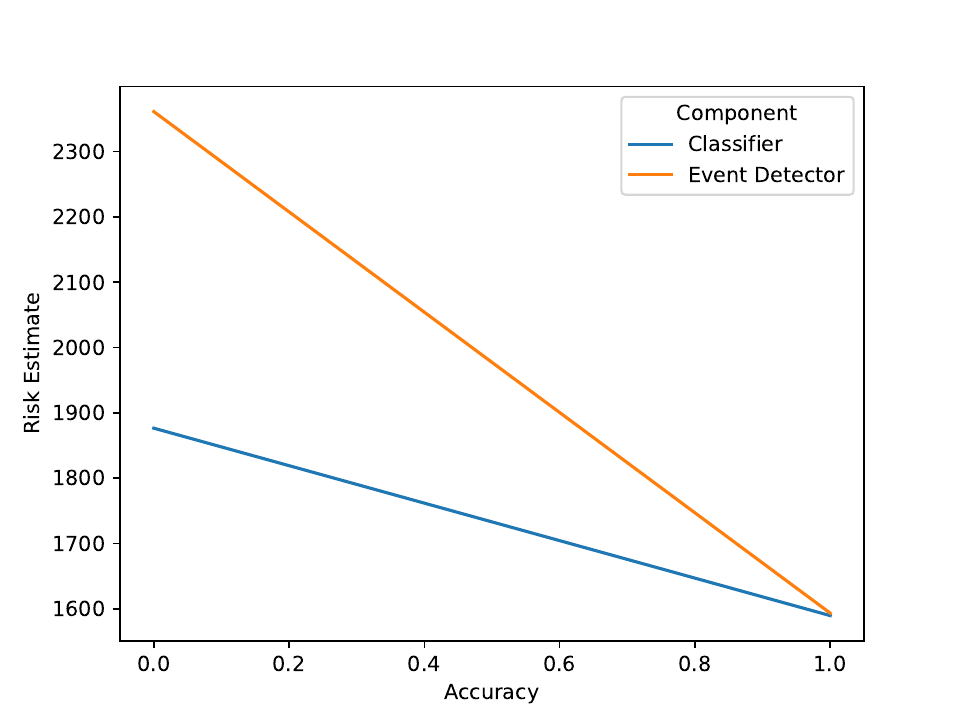}
    \caption{Plot showing the relationship between classifier/event detector accuracy and estimated system risk.}
    \label{fig:cba}
\end{figure}

Suppose, for instance, that we want to continue development such that the overall system risk is reduced by some margin. There are in this case two possible paths of action: the first is to improve classifier accuracy --- i.e. \(p(f(x)=y)\) --- and the second is to improve the event detector accuracy --- i.e. \(p(d_{e}(x)=E\)\footnote{It is worth noting that these may not necessarily be independent; an improved \gls{dnn} may result in \gls{ood} detectors that more readily discriminate \gls{ind} and \gls{ood} data. Improving the \gls{ood} detector will also improve the rate estimates, reducing error.}. By sweeping over a range of values for these two parameters, we can estimate the expected savings in terms of reduction of risk. This is shown in \Cref{fig:cba}. The results indicate that a given increase in \gls{ood} detector accuracy yields increased reduction in risk than an equivalent increase in classifier accuracy, and thus that this is the most effective use of development resources. In general, one can express the estimated savings associated with improving a parameter by taking the derivative of the tree with respect to said parameter. 

\section{Discussion and Limitations}\label{ch:discussion}

Overall, our results demonstrate that accounting for the incidence of adverse data events, notably distributional shifts, results in more credible and precise estimates of overall accuracy. Our approach affords several distinct advantages over conventional approaches to evaluation. In particular:
\begin{itemize}
    \item \textbf{Improved uncertainty quantification}. By continuously computing accuracy estimates for a trace of predictions, our approach is essentially serving as an  uncertainty quantifier and a form of \gls{rv}. The resulting estimates can be used to assess the credibility of the predictions at runtime. This provides increased transparency with respect to the systems level of performance, and can feasibly be used in conjunction with explanations to effectively communicate the behaviour of a given \gls{dnn}. 
    \item \textbf{Resilience to evolving data environments}. In contrast to competing approaches, our approach explicitly models any changes in the frequency of events by iteratively updating the event probability estimate at runtime. This way, the system remains adaptive to shifting data distributions, ensuring that accuracy estimates remain relevant even as the frequency of the events changes. Furthermore, by continuously integrating new observations, our method mitigates the risk of outdated or misleading performance assessments, which are common in static evaluation frameworks. This adaptability is particularly critical in safety-sensitive applications, where failure to recognize shifts in the data generation environment can lead to erroneous decision-making and increased operational risk.
    \item  \textbf{Risk assessment}. By associating nodes in the event-tree with a given cost, it is possible to compute more interpretable and informative metrics upon which value-judgments can be based. The associated costs can be tuned according to the requirements of the domain and adjusted in accordance with emerging information. These risk estimates concretely characterize the costs and benefits associated with a system, and can be be readily communicated to stakeholders, once again increasing trustworthiness. 
    \item \textbf{Extensibility and Modularity}. Our approach is is capable of accounting for adverse data events of any kind as long as a corresponding detector can be implemented. In a multimedia system, for instance, one may construct event-trees for each data stream, which can be updated and extended independently of one another. Moreover, while we have primarily considered accuracy- and risk-metrics, our approach can easily integrate other evaluation metrics.
\end{itemize} 

We envision that our approach may endow \glspl{dls} with an increased degree of trustworthiness, serve as a basis for ensuring regulatory compliance, and inform value-judgements with respect to their development and deployment. Ultimately, we believe that embedding risk-aware dynamic evaluation into \glspl{dls} will play a crucial role in fostering safer and more dependable AI-driven  systems.

Our work has several limitations that need to be taken into account. Firstly, the assumption of a static shift environment within each trace window, as noted in Section 3.2.1, represents another limitation of our current approach. While our experimental results show that the framework performs well across various datasets and shift frequencies, environments with highly dynamic shift patterns might benefit from more sophisticated temporal modeling. Specifically, in deployment scenarios where the distribution of data changes rapidly and unpredictably, the lag between environmental changes and their reflection in our risk estimates could lead to temporarily inaccurate assessments. Future work could explore adaptive trace length selection or more sophisticated temporal models to better capture these dynamics without sacrificing the computational efficiency of our current approach.

The efficacy of our method depends on a sufficiently performant event detector, e.g. a \gls{ood} detectors. While the relatively simple \gls{ood} detectors we studied in this work seemed to suffice to this end, recent work has shown that many \gls{ood} detectors do not necessarily generalize well across types and severities of shifts~\cite{loss_regression}. While we did not observe this on our benchmarks, it is worth noting that discrepancies in the detection rates across different kinds of \gls{ood} data will result in poor accuracy and risk estimates. 

Another limitation of our method is the fact that attaining low error accuracy estimates for \gls{ood} data requires an \gls{ood} calibration dataset that sufficiently represents the \gls{ood} accuracy in deployment. As is evident in \Cref{fig:classifier_errors}, different types of \gls{ood} data may exhibit significantly differing levels of accuracy. This will naturally lead to accuracy estimate errors that increase along with the shift frequency, as shown in \Cref{fig:synth_dsd_results} for the Polyp and Office31 datasets. One plausible solution to this is to leverage the expressive power that has been observed in many \gls{ood} detectors implementations~\cite{loss_regression} and distinguish between degrees of accuracy.That is, instead of having one branch for correct predictions and one for incorrect predictions, one can distinguish between binned intervals of expected accuracy, or, equivalently, loss. We plan on investigating this in further detail in future work. 

We have also neglected to consider execution requirements. The implementation of \gls{ood} detectors will necessarily introduce a degree of overhead, further complicated by the need to analyze the verdict traces as part of the processing performed by our method. For domains with real-time requirements, this may represent a significant obstacle. We do, however, contend that continued work on \gls{ood} detector algorithms and an efficiency-oriented implementation of our algorithms is likely to reduce this overhead considerably, and moreover that benefits of utilizing our approach may outweigh the costs to execution time. 

Throughout our work we have presented results for batched configurations. This is because, as discussed in \Cref{ch:methodology}, many systems typically operate on batches of data by default, and since \gls{ood} detector implementations often leverage this property towards increasing accuracy. There is a caveat to operating on batched data, however, and that is that batched \gls{ood} detectors implicitly assume approximate uniformity with respect to \gls{ood}-ness in the batches on which they operate. While this assumption often holds in video domains, it does not necessarily hold in other domains, e.g. in cloud systems or other systems for which there is not a bias with respect to the elements of a batch. Configurations with high batch sizes will in this case contain a mix of \gls{ind} and \gls{ood} samples, which typically result in poor performance as shown in \Cref{fig:nonuniform}. For a batch size of 1, this naturally does not pose a concern, but since low batch sizes are associated with lower \gls{ood} detector accuracies, it often also entail higher accuracy- and risk estimate errors, as evident for the NICO++, CCT, and Office31 datasets in \Cref{fig:dsd_results}. 

We have only considered a small sample of \gls{ood} detector implementations, primarily selected due to their popularity and relative ease of implementation. These methods have to a certain extent been made obsolete my more modern, state-of-the-art methods, however. Implementing such methods will likely result in further reduction in error rates \gls{ood} detectors. We plan to assess our approach on state-of-the-art approaches in future work. Nevertheless, even with these sub-optimal implementations, we attain generally favourable error rates. 

As described in \Cref{ch:framework}, our method is intended to be suitable for potentially any type of data event and any data modality. We have, however, solely considered distributional shifts in visual domains in this work. This is primarily due to the availability of \gls{ood} detector implementations, whereas for instance detectors for data containing sensitive qualities in the context of e.g. privacy or fairness have not been extensively studied for visual neural networks. While it is possible to conduct further evaluations of our methods in the context of for instance \glspl{llm}, with guardrails as the event detectors, this was not considered feasible due to a lack of computational resources, suitable datasets, etc. We do note, however, that applying our methods to \gls{llm} constitutes a pertinent and interesting avenue of future work. 

Finally, our methods only account for one type of event. In a production system, there may be several types of potentially affective events that may need to be accounted for, for instance, adversarial attacks, fairness violations, and jailbreaking. While it is possible to relatively simply implement several instances of our methods to account for this, such an approach assumes that these events are independent of one another, which may not be the case. Fairness violations may, for instance, arise as a result of distributional shift. Characterizing the inter-dependence of these events may be useful, and can conceivably be implemented by connecting nodes across separate event trees. We leave this as future work, largely due to the scarcity of suitable datasets to this end. 

\section{Conclusion}\label{ch:conclusion}
 In this work, we have outlined a method for probabilistic runtime verification, evaluation, and risk assessment of visual deep learning systems. We showed that by accounting for and modelling the incidence of distributional shifts, and the resulting probabilities that the corresponding neural network yields correct predictions, it is possible to compute credible estimates of overall system accuracy and risk at runtime, with average error margins typically ranging upwards of 0.1. We further demonstrated the practical utility of our methods in a polyp segmentation case study, therein how risk estimates can be used as a basis risk specification and cost-benefit analyses. Overall, we conjecture our work constitutes a meaningful initial step towards more credible evaluation of deep learning systems, and that continued work can improve the credibility of these systems in deployment scenarios.
 
\bibliographystyle{ACM-Reference-Format}
\bibliography{bibliography}

\newpage
\appendix
\section{Batch Uniformity}\label{a:bu}
\begin{figure*}[!t]
    \centering
    \includegraphics[width=\linewidth]{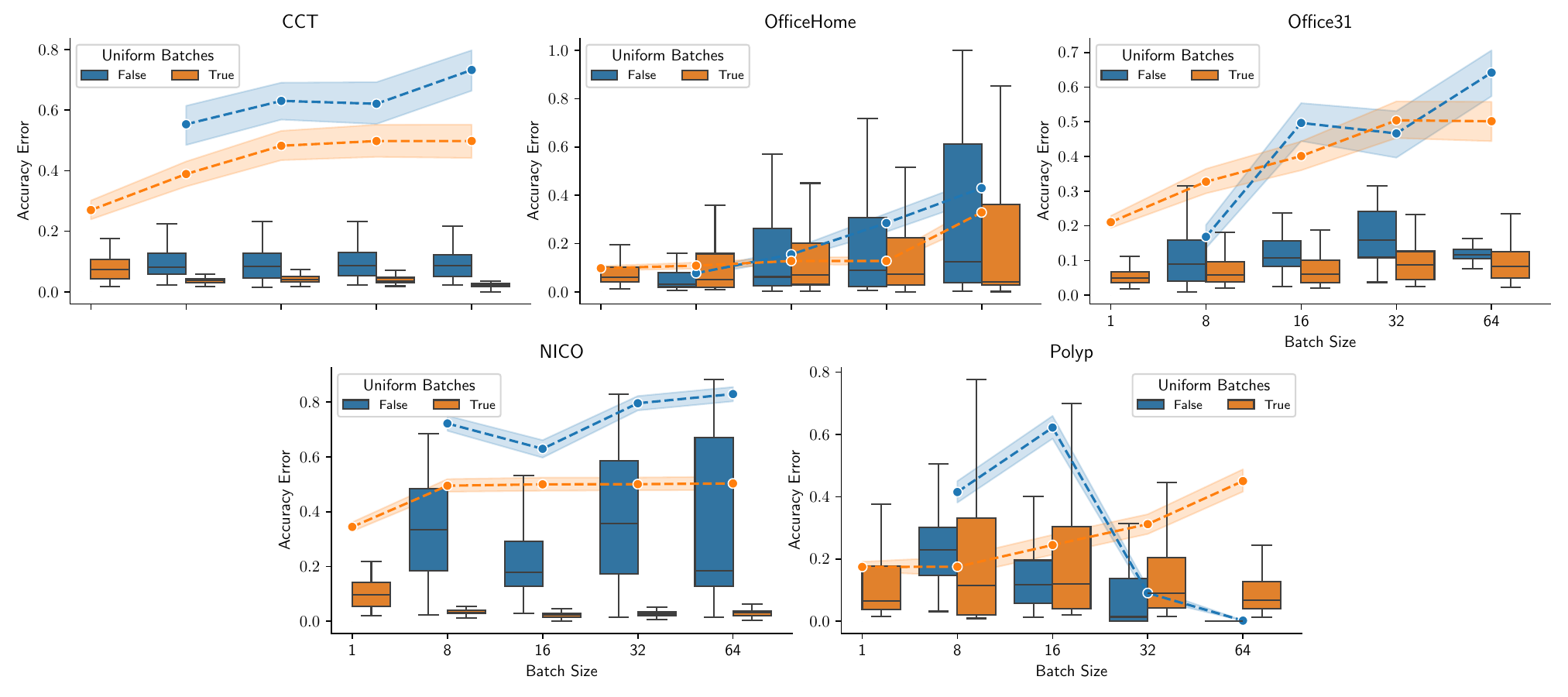}
    \caption{Accuracy estimate errors for uniform vs non-uniform batches. Note that a batch size of 1 outperforms nearly all configurations of higher batch size for non-uniform batches.}
    \label{fig:nonuniform}
\end{figure*}

In \Cref{fig:dsd_acc}, the results assume uniformity in the batches. While this is a relatively sound assumption for video domains and other contexts where incoming batches are sampled with bias, it is not necessarily always the case. We therefore include results for non-uniform batches in \Cref{fig:nonuniform}. Note that in this case, high batch sizes in fact results in larger errors than simply using a batch size of one. Thus, while batching indeed does improve \gls{ood} detector accuracy and may result in lower accuracy- and risk estimation errors as a result for uniform batches, it is generally not advisable to use a batched configuration if there is doubt about the degree of uniformity in the batches. 
\section{Batch Size Breakdown}\label{a:bsb}

For the purposes of presentation, the results in \Cref{fig:dsd_acc_errors_by_rate} were averaged over batch sizes. For completeness, we provide a breakdown of this plot across batch sizes. 
\begin{figure*}[!t]
    \centering
    \includegraphics[width=\linewidth]{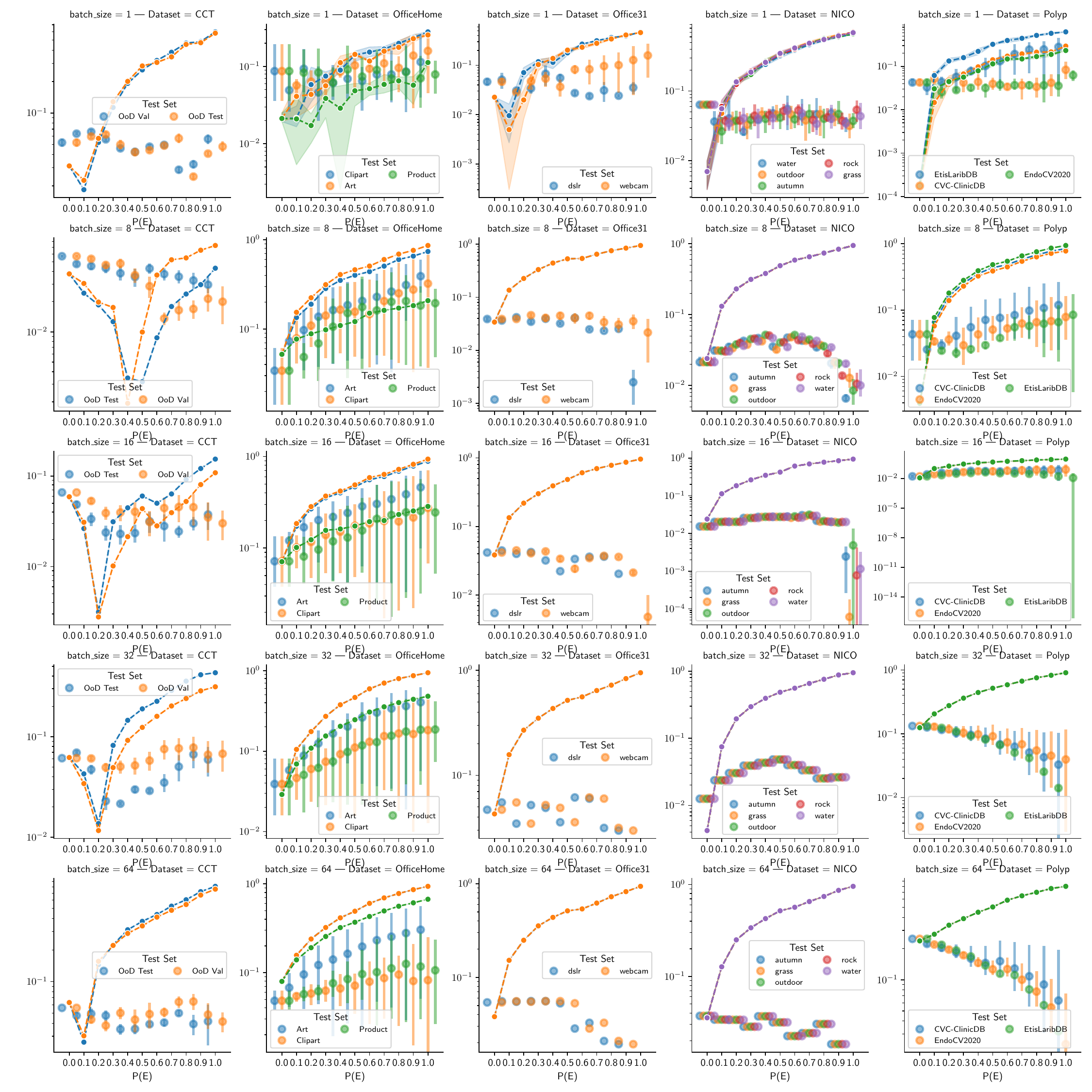}
    \caption{Accuracy estimate errors over all tested batch sizes and rates for the best performing OOD detector for each dataset.}
    \label{fig:bs_error_breakdown}
\end{figure*}

\end{document}

%% file: glossaries.tex
\newacronym{name}{DL-PRA}{Deep Learning Probabilistic Risk Assessment}
\newacronym{dnn}{DNN}{Deep Neural Network}
\newacronym{dls}{DLS}{Deep Learning System}
\newacronym{llm}{LLM}{Large Language Model}
\newacronym{ood}{OOD}{Out of Distribution}
\newacronym{gam}{GAM}{Generalized Additive Model}

\newacronym{ind}{InD}{In-Distribution}
\newacronym{dsd}{DSD}{Distributional Shift Detector}
\newacronym{ml}{ML}{Machine Learning}
\newacronym{rv}{RV}{Runtime Verification}
\newacronym{pra}{PRA}{Probabilistic Risk Assessment}

\newacronym{iid}{IID}{independently and identically distributed}
\newacronym{vae}{VAE}{Variational Autoencoder}

\newacronym{ks}{KS}{Kolmogorov-Smirnoff}
\newacronym{mw}{MW}{Mann-Whitney U}

\newacronym{fpr}{FPR}{False Positive Rate}
\newacronym{fnr}{FNR}{False Negative Rate}
\newacronym{tpr}{TPR}{True Positive Rate}
\newacronym{tnr}{TNR}{True Negative Rate}
\newacronym{ba}{BA}{Balanced Accuracy}
\newacronym{rmse}{RMSE}{Root Mean Square Error}
\newacronym{mape}{MAPE}{Mean Average Percentage Error}
\newacronym{mae}{MAE}{Mean Average Error}

%% file: tables/benchmarking_datasets.tex
\begin{tabularx}{\linewidth}{l p{1cm} p{1.1cm} p{1cm} p{1.1cm} X}
            \toprule
\rotatebox{60}{Dataset}                 & \rotatebox{60}{No. Images}  & \rotatebox{60}{Resolution} & \rotatebox{60}{No. Classes} & \rotatebox{60}{Train/Test Domains} & \rotatebox{60}{Description}\\
\midrule
\textbf{NICO++}         & 230000    & Varying   & 60    & 1/5   &  domain-adaptation classification dataset partitioned by a number of contexts. \\
\textbf{Office-Home}    & 15500     & Varying         & 65    & 1/3   &  A domain-adaptation classification dataset partitioned by modality\\
\textbf{Office-31}      & 4110      & Varying   & 31    & 1/2   & A domain-adaptation classification dataset partitioned by camera type\\
\textbf{CCT}            & 243187    & 1024X747         & 15    & 65/65 & a domain-adaptation/generalization benchmark consisting of images taken from different camera-traps. \\
\textbf{Polyps}         & 1323         & Varying         & 1    & 2/1   & Three colonoscopy datasets, Kvasir-SEG~\cite{kvasir}, ENDOCV2020~\cite{endocv2020}, and Etis-LaribDB\cite{etis-larib}, each consisting of colonoscopic images of polyps taken from differing centers and with different equipment. Note that in contrast to the previous datasets, these are segmentation datasets.\\
                 \bottomrule
\end{tabularx}

%% file: tables/ood_detectors.tex
 \begin{tabularx}{\linewidth}{XX}
    \toprule
         Detector & Description \\
    \midrule
         \textbf{Entropy}~\cite{dsd_baseline} \\ 
         \hspace{1em}$H(x) = - \sum_{y} p_\theta(y|x) \log p_\theta(y|x)$ &  
         Predictive entropy of the model output distribution. \\[0.75em]
         
         \textbf{GradNorm}~\cite{gradients_distshift} \\ 
         \hspace{1em}$||\nabla_{\theta}\, H(x)||_{2}$ &  
         $l_2$ norm of the gradient of entropy w.r.t.\ model parameters. \\[0.75em]
         
         \textbf{Energy}~\cite{dsd_energy} \\ 
         \hspace{1em}$E(x) = -\log \sum_{y} e^{f_\theta(x)_y}$ &  
         Energy score, i.e.\ negative log-sum-exp of logits $f_\theta(x)$. \\[0.75em]
         
         \textbf{kNN}~\cite{knn_dsd} \\ 
         \hspace{1em}$\min_{x_i \in \mathcal{D}_{\text{train}}} ||z(x) - z(x_i)||_2$ &  

         Distance between test encoding $z(x)$ and its nearest neighbour $z_k(x)$ in the training set. Note that the original paper extends this to multiple k values. \\[0.75em]
         
         \textbf{MSP}~\cite{dsd_baseline} \\ 
         \hspace{1em}$\max(p_\theta(y|x))$ &  
         Maximum softmax score. \\[0.75em]
         
         \textbf{Typicality}~\cite{typicality_dsd} \\ 
         \hspace{1em}$-\log p_{\text{GLoW}}(x)$ &  
         Negative log-likelihood under the GLoW generative model. \\
    \bottomrule
    \end{tabularx}

%% file: tables/ood_detector_accuracy.tex
\begin{tabular}{lccccc}
\toprule
\multicolumn{6}{c}{Organic} \\
\midrule
Feature &        CCT &       NICO &   Office31 & OfficeHome &      Polyp \\
\midrule
Energy           &  0.69 &  0.72 &  0.57 &  0.51 &  0.74 \\
Entropy          &  0.63 &  0.84 &  0.62 &  0.51 &  0.59 \\
GradNorm         &  0.53 &  0.82 &  0.60 &  \textbf{0.61} &  0.57 \\
Softmax          &  0.69 &  \textbf{0.90} &  0.63 &  0.52 &   -   \\
Typicality       &  0.57 &  0.52 &  0.89 &  0.52 &  0.55 \\
kNN              &  \textbf{0.89} &  0.70 &  \textbf{0.88} &  0.54 &  \textbf{0.75}   \\
\midrule
\bottomrule
\end{tabular}

%% file: tables/training_configuration.tex
\begin{tabular}{ll}
\toprule
     \textbf{Component} & \textbf{Choice} \\
\midrule
     Framework & Pytorch Lightning~\cite{lightning}.\\
     Optimizer & Adam \\
     Scheduler & Cosine Annealing w/ Warm Restarts\\
     Batch size & 16 \\
     Data augmentation & Random flip, Random rotate\\
     Initial learning Rate & 1e-3 \\
     Model Selection & Early Stopping \\
     Loss & Cross Entropy Loss/Jaccard Loss \\
\bottomrule
\end{tabular}

%% file: tables/classifier_accuracy.tex
 \begin{tabularx}{\linewidth}{XXXXX}  
        \begin{subtable}[t]{0.5\linewidth}
            \centering
            \begin{tabular}{lc}
            \toprule
                \multicolumn{2}{c}{\textbf{CCT}} \\
                \midrule
                \textbf{Train} & 0.95 \\
                \textbf{InD Val.} & 0.87 \\
                \textbf{InD Test} & 0.91 \\
                \textbf{OoD Val} & 0.31 \\
                \textbf{OoD Test} & 0.31 \\
                \bottomrule
            \end{tabular}
        \end{subtable} &
        \begin{subtable}[t]{0.5\linewidth}
            \centering
            \begin{tabular}{lc}
               \toprule
                \multicolumn{2}{c}{\textbf{NICO++}} \\
                \midrule
                \textbf{Train} & 0.93 \\
                \textbf{InD Val.} & 0.93 \\
                \textbf{InD Test} & 0.94 \\
                \textbf{Autumn} & 0.25 \\
                \textbf{Grass} & 0.24 \\
                \textbf{Outdoor} & 0.26 \\
                \textbf{Rock} & 0.24 \\
                \textbf{Water} & 0.27 \\
                \bottomrule
            \end{tabular}
        \end{subtable} &
        \begin{subtable}[t]{0.5\linewidth}
            \centering
            \begin{tabular}{lc} 
                \toprule
                \multicolumn{2}{c}{\textbf{Office31}} \\
                \midrule
                \textbf{Train} & 0.50 \\
                \textbf{InD Val.} & 0.43 \\
                \textbf{InD Test} & 0.39 \\
                \textbf{Dslr} & 0.05 \\
                \textbf{Webcam} & 0.05 \\
                \bottomrule
            \end{tabular}
        \end{subtable} &
        \begin{subtable}[t]{0.5\linewidth}
            \centering
            \begin{tabular}{lc}  
                \toprule
                \multicolumn{2}{c}{\textbf{OfficeHome}} \\
                \midrule
                \textbf{Train} & 0.98 \\
                \textbf{InD Val.} & 0.39 \\
                \textbf{InD Test} & 0.40 \\
                \textbf{Art} & 0.13 \\
                \textbf{Clipart} & 0.11 \\
                \textbf{Product} & 0.30 \\
                \bottomrule
            \end{tabular}
        \end{subtable} &
        \begin{subtable}[t]{0.5\linewidth}
            \centering
            \begin{tabular}{lc}  
                \toprule
                \multicolumn{2}{c}{\textbf{Polyp}} \\
                \midrule
                \textbf{Train} & 0.96 \\
                \textbf{InD Val.} & 0.90 \\
                \textbf{InD Test} & 0.87 \\
                \textbf{CVC-Clinic} & 0.71 \\
                \textbf{EndoCV2020} & 0.74 \\
                \textbf{EtisLarib} & 0.33 \\
                \bottomrule
            \end{tabular}
        \end{subtable}
    \end{tabularx}